\documentclass[letterpaper, 10 pt, conference]{ieeeconf}  

\IEEEoverridecommandlockouts                              

\usepackage{amsmath}
\usepackage{amsthm}
\usepackage{algorithm}
\usepackage[noend]{algpseudocode}
\usepackage{amssymb}  
\usepackage[dvipsnames]{xcolor}
\usepackage{graphicx}
\usepackage{booktabs}
\usepackage{hyperref}
\usepackage{mathtools}
\usepackage{multirow}
\usepackage{siunitx}
\usepackage{cite}

\DeclareMathOperator*{\minimize}{minimize}				
\DeclareMathOperator*{\st}{subject\,to}					

\newcommand{\Xs}[0]{\ensuremath{\mathcal{X}}}						
\newcommand{\Us}[0]{\ensuremath{\mathcal{U}}}						
\newcommand{\Vs}[0]{\ensuremath{\mathcal{V}_M}}						
\newcommand{\Vshat}[0]{\ensuremath{\hat{\mathcal{V}}_M}}			

\newcommand{\Expect}{{\rm I\kern-.3em E}}				

\newtheorem{assumption}{Assumption}

\makeatletter
\renewenvironment{proof}[1][\proofname]{\par
\pushQED{\qed}%
\normalfont \topsep6\p@\@plus6\p@\relax
\trivlist
\item\relax
{\itshape
#1\@addpunct{.}}\hspace\labelsep\ignorespaces
}{%
\popQED\endtrivlist\@endpefalse
}
\makeatother
\graphicspath{{figures/}}

\algnewcommand\And{\textbf{and} }
\algnewcommand\Or{\textbf{or} }

\algrenewcommand\algorithmicforall{\textbf{parallel-for}}

\title{\LARGE \bf
Parallel-Constraint Model Predictive Control: Exploiting Parallel Computation for Improving Safety
}

\author{Elias Fontanari$^{1}$, Gianni Lunardi$^{1}$, Matteo Saveriano$^{1}$ and Andrea Del Prete$^{1}$
\thanks{$^{1}$The authors are with the Industrial Engineering Department, University of Trento, Via Sommarive 11, 38123, Trento, Italy. {\tt \{name.surname\}@unitn.it}}%
\thanks{
Financed by the European Union - Next Generation EU, Mission 4 Component 2 - CUP E53D23001130006 (STARLIT), CUP E53D23001020001 (ARIEL), and project INVERSE (Grant Agreement n. 101136067).}
\thanks{© 2025 IEEE. Personal use of this material is permitted. Permission
from IEEE must be obtained for all other uses, in any current or future
media, including reprinting/republishing this material for advertising or
promotional purposes, creating new collective works, for resale or
redistribution to servers or lists, or reuse of any copyrighted
component of this work in other works.}
}

\begin{document}

\maketitle

\thispagestyle{empty}
\pagestyle{empty}

\begin{abstract}
Ensuring constraint satisfaction is a key requirement for safety-critical systems, which include most robotic platforms.
For example, constraints can be used for modeling joint position/velocity/torque limits and collision avoidance.
Constrained systems are often controlled using Model Predictive Control, because of its ability to naturally handle constraints, relying on numerical optimization.
However, ensuring constraint satisfaction is challenging for nonlinear systems/constraints. 
A well-known tool to make controllers safe is the so-called \emph{control-invariant set} (a.k.a. safe set).
In our previous work, we have shown that safety can be improved by letting the safe-set constraint recede along the MPC horizon.
In this paper, we push that idea further by exploiting parallel computation to improve safety. 
We solve several MPC problems at the same time, where each problem instantiates the safe-set constraint at a different time step along the horizon. 
Finally, the controller can select the best solution according to some user-defined criteria. 
We validated this idea through extensive simulations with a 3-joint robotic arm, showing that significant improvements can be achieved in terms of safety and performance, even using as little as 4 computational cores.
\end{abstract}

\section{Introduction}\label{sec:introduction}
Guaranteeing safety is a fundamental requirement in almost all robotics applications. 
Safety is typically formulated via a set of equality/inequality constraints that the system should satisfy at all times. 
For instance, such constraints can model joint limits, actuation bounds, and collision avoidance.
While formulating these constraints is typically straightforward, ensuring their persistent satisfaction is extremely challenging. 

This is the case for both recent data-driven approaches, often relying on Reinforcement Learning (RL) algorithms, and for model-based control methods such as Model Predictive Control (MPC). 
Indeed, if the system dynamics is nonlinear, MPC methods in general cannot easily guarantee safety. 
The most common approach for guaranteeing safety relies on the knowledge of a so-called \emph{safe set} (a.k.a. control-invariant set)~\cite{Blanchini1999, Grune2017}, or, equivalently, a Control Barrier Function (CBF)~\cite{Ames2014, Wu2019}. 

Nonetheless, computing exact safe sets for nonlinear systems is generally unfeasible. Hence, professionals have relied on numerical methods that can generate approximations of such sets/functions~\cite{Djeridane2006, Coquelin2007,Jiang2016,Rubies2016,Hsu2021,Dawson2023, Zhou2020, LaRocca2023}.
Unfortunately, safety guarantees are compromised if the safe set is not exact. 
A more in-depth discussion of MPC methods that can guarantee safety is postponed to Section~\ref{sec:preliminaries}.

An alternative to MPC is \textit{Model Predictive Shielding} (MPS)~\cite{Li2020a, Bastani2021}.
This approach assumes the knowledge of a backup policy that can drive the state to an invariant set.
Contrary to the invariant set required by MPC methods, this invariant set can be potentially very small, without negatively affecting the basin of attraction of the controller. 
However, finding a backup policy and the associated (small) invariant set has a similar complexity to finding a (large) safe set.

Despite the abundant literature on safe control methods, little work has focused on exploiting parallel computing for improving safety. 
MPC optimization problems can be solved exploiting parallel computation, either on CPU, GPU, or FPGA~\cite{Abughalieh2019, Adabag2023, jeon2024cusadi}, but optimization algorithms are hard to parallelize and therefore only limited speed-ups can be achieved.
Moreover, solving the MPC faster does not directly improve safety, as it simply reduces the time delay due to the computation.

Sampling-based MPC methods~\cite{Rajamaki2016,Williams2017,howell2022predictive} are natural candidates for parallel computation, given the simple parallelization of their algorithms.
However, they do not address directly the safety aspect.
The only exception is Statistical MPS, which relies on forward simulating the backup policy considering different instances of the system noise~\cite{Bastani2021}. 
While the authors did not discuss this aspect, it is clear that such operation can easily exploit parallel computation.
However, this only makes sense when the system is subject to noise and a statistical certification is acceptable.

Our main contribution is a novel MPC scheme that exploits parallel computation to directly improve safety. 
This is fundamentally different from previous work exploiting parallel computation for MPC, because we do not merely try to solve the problem faster, but instead try to achieve a \emph{better} solution (in terms of safety), in the same amount of time.
Our method can be seen as an extension of our previous work on \emph{Receding-Constraint MPC}~\cite{Lunardi2024} (summarized in Section~\ref{sec:receding_constraint}), where we showed that by making the safe-set constraint recede along the horizon we could achieve superior theoretical and practical safety than classic terminal-constraint approaches.
With \emph{Parallel-Constraint MPC} (described in Section~\ref{sec:parallel_mpc}) we push this idea further. 
Rather than solving a single MPC problem, we solve several MPC problems, each one constraining the state inside the safe set at a different time step along the horizon. 
This enhances our chances of getting a feasible solution, therefore improving safety.
Moreover, among all solved problems, we can choose the one that satisfies the safe-set constraint the furthest along the horizon, therefore enlarging our safety window.
Our simulation results with a 3-degree-of-freedom robot manipulator, reported in Section~\ref{sec:results}, show that \emph{Parallel-Constraint} MPC can significantly outperform \emph{Receding-Constraint} MPC, completing up to $21\%$ more tasks while failing up to $6\%$ less tasks.
Significant improvements were also achieved when using a limited number of computational units (between 4 and 16).

\section{Preliminaries}
\label{sec:preliminaries}
This section quickly reviews a few relevant concepts from the literature: backward-reachable sets, control-invariant sets, and Recursive Feasibility in MPC.

\subsection{Notation}
\begin{itemize}
    \item $\mathbb{N}$ denotes the set of natural numbers;
    \item $\{ x_i \}_0^N$ and $X$ denote a discrete-time trajectory given by the sequence $(x_0, \dots, x_N)$;
    \item $x_{i|k}$ denotes the state at time step $k+i$ predicted when solving the MPC problem at time step $k$;
\end{itemize}

\subsection{Problem statement}
Let us consider a discrete-time dynamical system with state and control constraints:
\begin{equation} \label{eq:sys_dynamics}
    x_{i+1} = f(x_i, u_i),
\qquad
    x \in \Xs, \qquad u \in \Us .
\end{equation}
The objective is to design a controller that ensures \emph{safety} (i.e., constraint satisfaction), while minimizing a user-defined cost function.
We define $\mathcal{S}$ as the subset of \Xs\ containing all the equilibrium states:
\begin{equation}
    \mathcal{S} = \{ x \in \Xs\ | \, \exists \, u \in \Us : x = f(x,u) \}.
\end{equation}
To ensure safety, we will rely on the \emph{M-Step Backward-Reachable Set}~\cite{Blanchini1999} of $\mathcal{S}$, which we denote as $\Vs$. 
Mathematically, it is defined as the subset of \Xs\ starting from which it is possible to reach $\mathcal{S}$ in $M$ steps:
\begin{equation}
\begin{aligned}
\Vs \triangleq \{ x_0 \in \Xs \, | \, \exists \{u_i\}_{0}^{M-1}: \, \,
& x_M \in \mathcal{S}, x_i \in \Xs, \\
& u_i \in \Us, \forall \, i=0,\dots,M-1 \} .
\end{aligned}
\end{equation}
Being a backward reachable set of equilibrium states, the set $\Vs$ is control-invariant~\cite{Blanchini1999}.
Namely, starting inside $\Vs$, we can remain inside $\Vs$ forever.
Knowledge of the set $\Vs$ would make it easy to design a safe controller. 
However, we cannot assume to know \Vs\ in general, because its computation can be extremely complex. 
We rely instead on the following, more realistic, assumption.
\begin{assumption} \label{ass:conservative_set}
    We know a conservative approximation of the set $\Vs$: 
    \begin{equation}
        \Vshat \subseteq \Vs
    \end{equation}
    Note that $\Vshat$ need not be control invariant in general.
\end{assumption}
As discussed above, different methods exist to compute numerical approximations of $\Vs$. We have chosen to use a slightly modified version of the Viability-Boundary Optimal Control (VBOC) method~\cite{LaRocca2023}. The resulting approximation of \Vs\ can be easily made conservative by introducing a user-defined safety margin, which we explain in more detail in Section~\ref{sec:results}. 
In the following, we summarize standard MPC approaches to exploit $\Vshat$ for achieving safety.

\subsection{Model Predictive Control and Recursive Feasibility}
\label{ssec:recursive_feasibility}
Let us consider the following MPC problem:
\begin{subequations}
\begin{align}
        \minimize_{X,U}  &\quad \sum_{i = 0}^{N-1} \ell_i(x_i,u_i) + \ell_N(x_N)  \label{eq:mpc_cost} \\
        \st &\quad x_0 = x_{init} \label{eq:mpc_initial_conditions}\\
            &\quad x_{i+1} = f(x_i, u_i) \qquad i = 0 \dots N-1 \label{eq:mpc_dynamics}\\
            &\quad x_i \in \Xs, u_i \in \Us \qquad \;\; i = 0 \dots N-1 \label{eq:mpc_path_constraints}\\
            &\quad x_{N} \in \mathcal{X}_N \label{eq:mpc_terminal_constraint}
\end{align}
\label{eq:mpc}
\end{subequations}%
where $\ell(\cdot)$/$\ell_N(\cdot)$ is the running/terminal cost, $x_{init}$ is the current state, and $\mathcal{X}_N \subseteq \Xs$ is the terminal set~\cite{Mayne2000}.

Even if MPC is probably the most commonly used approach for controlling constrained systems, ensuring constraint satisfaction remains difficult when the dynamics or the constraints are nonlinear.
A well-known approach to ensure safety is to rely on Recursive Feasibility (RF).
RF guarantees that, if an MPC problem is feasible at the first loop, it remains feasible forever.


RF can be guaranteed by choosing the terminal set $\mathcal{X}_N$ to be a \emph{control-invariant} set. 
While theoretically sound, the practical problem of this approach is that control-invariant sets are in general extremely challenging to compute for nonlinear systems/constraints.
%


Since in general we only know $\Vshat$, which may not be control invariant, using it as terminal set does not ensure RF.
In other words, the MPC problem could become unfeasible.
To deal with unfeasibility, one can relax the terminal constraint using a slack variable, which is heavily penalized in the cost function~\cite{Kerrigan2000soft, Zeilinger2014}.
This can temporarily recover feasibility, in the hope that perhaps the terminal constraint be satisfied again in the future. 
However, with this approach we cannot ensure \emph{safety}, nor RF.
Indeed, the slack variable allows the state to leave $\Vshat$, which eventually can lead to leaving \Xs.

\section{Receding-Constraint MPC with Safe Abort}
\label{sec:receding_constraint}
In our previous work~\cite{Lunardi2024}, we have introduced a novel MPC scheme that relies on two components: a Receding-Constraint MPC formulation and a safe task-abortion strategy.
We summarize them in the following, in a slightly generalized form.

\subsection{Receding-Constraint MPC} 
\label{ssec:receding_constraint}

The key idea of this MPC scheme is to adapt online the time step at which the state is constrained in $\Vshat$, making it recede along the horizon.
If at MPC loop $k-1$ we had $x_{r|k-1} \in \Vshat$, at loop $k$ we can be sure that it is possible to have $x_{r-1|k} \in \Vshat$ (assuming no disturbances and modeling errors).
Thus the constraint can be established in a hard way.
This guarantees safety for $r$ loops, where this \emph{receding constraint} slides backward along the horizon.
However, once it reaches time step 0, we can no longer rely on it to ensure safety.
Therefore, we maintain also a soft constraint to encourage the terminal state to be in $\Vshat$, leading to this MPC formulation:
\begin{equation}
    \begin{aligned}
        \minimize_{X,U, s}  &\quad \sum_{i = 0}^{N-1} \ell_i(x_i,u_i) + \ell_N(x_N) + w_s |s| \\
        \st &\quad \eqref{eq:mpc_initial_conditions}, \eqref{eq:mpc_dynamics}, \eqref{eq:mpc_path_constraints} \\
            & \quad x_r \in \Vshat, \, x_{N} \in \Vshat \oplus s. \\
    \end{aligned}
    \label{eq:receding_constr_mpc}
\end{equation}
After solving the MPC at loop $k-1$, we check whether $x_{N|k-1} \in \Vshat$; if so, at loop $k$ we move the receding constraint forward on $x_{N-1|k}$. 
Actually, we can even check whether $x_{i} \in \Vshat$, for any $i>r$, and if that is the case we can set $r=i-1$ at the next loop.
Otherwise, we let $r$ decrease. 
If we reach $r=1+L$ (with $L$ being a small non-negative user-defined integer, whose role is clarified in the following) then we trigger the safe task abort procedure.

\subsection{Safe Task Abortion} \label{ssec:task_abortion}
Our key idea to ensure safety relies on Assumption~\ref{ass:conservative_set} and on the following two assumptions.
\begin{assumption} \label{ass:computational_units}
We have access to two computational units: unit A and unit B. 
\end{assumption}
\begin{assumption} \label{ass:ocp_time}
We can solve the following Optimal Control Problem (OCP) for any $x_{init} \in \Vshat$, in $L+1$ time steps (with $0\le L \le N-1$):
\begin{equation}
    \begin{aligned}
        \minimize_{\{x_i\}_0^{M},\{u_i\}_0^{M-1}}  &\quad \sum_{i = 0}^{M-1} \ell_i(x_i,u_i)  + \ell_{M}(x_{M})\\
        \st & \quad \eqref{eq:mpc_initial_conditions}, \eqref{eq:mpc_dynamics}, \eqref{eq:mpc_path_constraints}, \, x_{M} = x_{M-1}
    \end{aligned}
    \label{eq:safe_abort_ocp}
\end{equation}
The choice of the cost function is irrelevant.
\end{assumption}
OCP~\eqref{eq:safe_abort_ocp} finds a trajectory reaching an equilibrium state from $x_{init}$.
Suppose we trigger the safe abort procedure at MPC loop $k$.
The procedure consists of these steps:
\begin{enumerate}
    \item unit A uses the MPC solution computed at loop $k-1$ to reach the safe state $x_{L+1|k-1} \in \Vshat$;
    \item in parallel, unit B solves  OCP~\eqref{eq:safe_abort_ocp}, using $x_{L+1|k-1}$ as initial state;
    \item after reaching $x_{L+1|k-1}$, we follow the solution of OCP~\eqref{eq:safe_abort_ocp} to reach an equilibrium state.
\end{enumerate}
The hyper-parameter $L$ should be set by the user as small as possible, while satisfying Assumption~\ref{ass:ocp_time}.

\subsection{Theoretical Properties}
In~\cite{Lunardi2024}, we have shown that the Receding-Constraint MPC scheme, coupled with the safe abort procedure, enjoys two interesting theoretical properties. 

The safe abort procedure ensures \emph{safety} as long as Assumptions~\ref{ass:conservative_set}, \ref{ass:computational_units} and~\ref{ass:ocp_time} are satisfied.
Moreover, Receding-Constraint MPC ensures \emph{recursive feasibility} under the assumption that $\Vshat$ is $(N-L)$-step control invariant. This means that, from any state inside $\Vshat$, it must be possible to reach a state in $\Vshat$ in at most $N-L$ time steps. 
This property is clearly weaker than the classic control invariance that is required by terminal constraint approaches.

\section{Parallel-Constraint Model Predictive Control}
\label{sec:parallel_mpc}
Instead of using a single state along the horizon to ensure safety, we could try to reach the safe set at different time steps exploiting parallel computation. 
As long as at least one state $x_p \in \Vshat$ (with $1 \le p \le N$), we can ensure constraint satisfaction because $x_1 \in \mathcal{V}_{M+p-1}$.
This is because from $x_1$ we can reach $x_p$ in $p-1$ steps. 
Ideally, we would like to include the following constraint in our problem formulation:
\begin{equation} \label{eq:or_constraints}
    (x_1 \in \Vshat) \, \lor \, (x_2 \in \Vshat) \lor \, \dots \, \lor \, (x_N \in \Vshat).
\end{equation}
However, this type of constraint (OR constraints) is extremely hard to deal with for numerical solvers.

\begin{figure}[tp]
    \centering
    \includegraphics[width=\columnwidth]{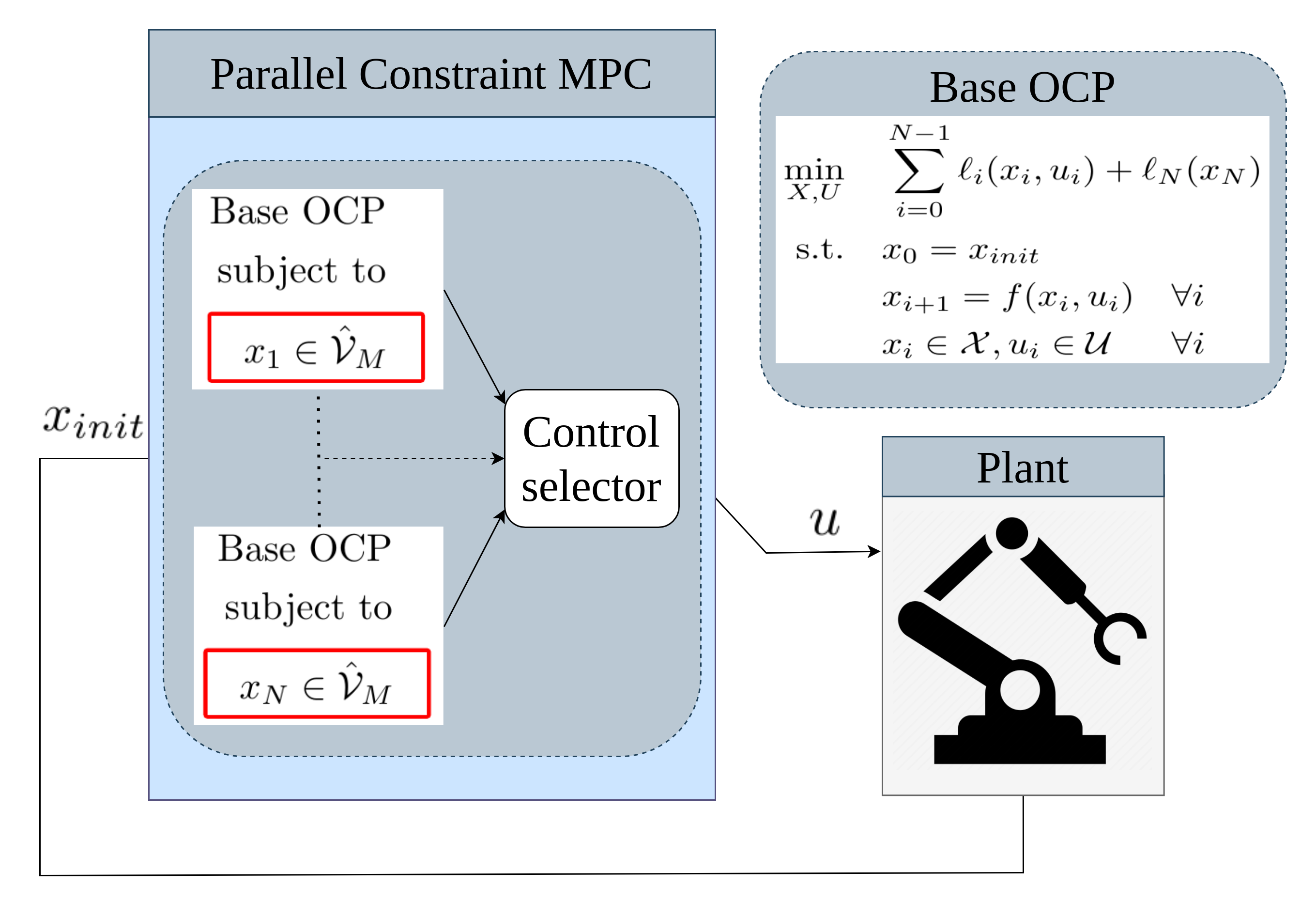}
    \caption{Parallel MPC scheme. At each control step, N different problems with form \eqref{eq:parallel_constr_mpc} are solved in parallel, then the best solution in term of safety is chosen and applied to the plant.}
    \label{fig:ParallelMpcSCHEME}
\end{figure}

If we have access to $N$ computational units, we can solve $N$ problems in parallel, each constraining the state inside \Vshat\ at a different time step $p \in [1, N]$:
\begin{equation}
    \begin{aligned}
        \minimize_{X,U}  &\quad \sum_{i = 0}^{N-1} \ell_i(x_i,u_i) + \ell_N(x_N)\\
        \st &\quad \eqref{eq:mpc_initial_conditions}, \eqref{eq:mpc_dynamics}, \eqref{eq:mpc_path_constraints} \\
            & \quad x_p \in \Vshat
    \end{aligned}
    \label{eq:parallel_constr_mpc}
\end{equation}
Then, we can use the solution of one of the problems that have been solved.
Since our main concern is safety, we decide to use the solution satisfying the safe set constraint at the furthest time.
A scheme of the Parallel Constraint MPC is depicted in Fig.~\ref{fig:ParallelMpcSCHEME} and a pseudo-code of the controller is available in Alg.~\ref{alg:parallel_mpc}.

\begin{algorithm}[t]
\caption{Parallel-Constraint MPC with Task Abortion}
\small
\begin{algorithmic}[1]
\Require 
Number of time steps $T$,
Initial state $x_0$,
OCP~\eqref{eq:parallel_constr_mpc},
Safe-abort OCP~\eqref{eq:safe_abort_ocp},
Safe-abort trigger time~$L$
    \State $r \leftarrow N$  \Comment{Safe horizon}
    \For{$t = 0 \rightarrow T-1$}
        \ForAll{$p \in [1, N]$}
            \State $X_p, U_p \leftarrow \textsc{OCP}(p, x_t)$ 
            \State $r_p \leftarrow$ NaN
            \For{$k = r \rightarrow N$} \Comment{Search last state in $\Vshat$}
                \If{$X_p[k] \in \Vshat$}
                    \State $r_p \leftarrow k$
                \EndIf
            \EndFor
        \EndFor
        \State $p \leftarrow \Call{ArgMax}{\{r_p\}_1^N}$ \Comment{Find best solution}
        \If{$p$ is NaN} \Comment{Use previous solution}
            \State $r \leftarrow r - 1$
            \State $X^*[:N-1], U^*[:N-2] \gets X^*[1:], U^*[1:]$
        \Else
            \State $r \leftarrow r_p - 1$
            \State $X^*, U^* \gets X_p, U_p$
        \EndIf
        \If{$r = L$} \Comment{Trigger Safe Abort}
            \State $\textsc{TriggerSafeAbort}(x_{L}^*)$ \Comment{Solve~\eqref{eq:safe_abort_ocp} in Unit B and then follow the computed trajectory}
        \EndIf
        \State $x_{t+1} \leftarrow f(x_t,u_0^*)$ \Comment{Simulate dynamics}
    \EndFor
\end{algorithmic}
\label{alg:parallel_mpc}
\end{algorithm}

\subsection{Limited Computational Units}
\label{ssec:limited_computation}
If less than $N$ computational cores are available, we should decide on which time steps to try to satisfy the safe set constraint. 
First of all, we decided to always allocate one computational unit to solve an OCP having the safe-set constraint at time step $r-1$, where $r$ is the time step at which the safe set constraint was satisfied at the previous MPC loop.
This is because, in theory, the safe-set constraint is feasible at that time step.

For allocating the remaining computational units, we have devised three possible strategies to choose the time steps.
\begin{itemize}
\item \emph{Uniform}: Approximately equally spaced along the horizon, considering that time step $r-1$ must always be included in the list. For instance, if we had 4 computational units, a horizon $N=20$, and $r=9$, we would choose the time steps 4, 8, 14, and 20. 
\item \emph{High}: The furthest down the horizon (i.e., $N, N-1, N-2, \dots$), which would provide safety for longer.
\item \emph{Closest}: The time steps associated to the smallest violations of the safe set constraint at the previous loop.
\end{itemize}

\subsection{Checking Constraint Satisfaction}
Our MPC implementation relies on the Real-Time Iteration (RTI) scheme~\cite{diehl2005nominal}, which boils down to performing a single iteration of Newton's method, without line search.
Therefore, after each solver iteration we need to check if constraints are satisfied.
In particular, also the system dynamics can be violated. This implies that, even if the predicted state trajectory satisfies all constraints, the real state trajectory obtained by forward integration of the predicted control inputs may violate some constraints. 
For this reason, for every problem, we need to perform a forward integration of the computed control trajectory, and check the constraints on the resulting state trajectory.


\subsection{Warm Start}
As typically done in MPC, we provide an initial guess to warm-start the solver. 
However, rather than providing a shifted version of the previous solution, we use a different approach, which we have observed to improve the results.
As mentioned above, the RTI scheme can result in violation of the dynamics constraints.
To help the solver satisfy the dynamics constraints, we provide an initial guess that satisfies them. 
This guess is obtained by forward integration of the predicted control trajectory.

\subsection{Theoretical Guarantees}
From a theoretical standpoint, \emph{Parallel-Constraint} and \emph{Receding-Constraint} MPC share similar properties. 
Indeed, both ensure \emph{recursive feasibility} under the assumption that $\Vshat$ is $(N-L)$-step control invariant, and both ensure \emph{safety} using the task-abortion strategy, under Assumptions~\ref{ass:conservative_set}, \ref{ass:computational_units}, and~\ref{ass:ocp_time}.

The variants of \emph{Parallel-Constraint} MPC that use a limited number of computational units maintain safety, but they may lose recursive feasibility, depending on how they are implemented.
In particular, one way to ensure recursive feasibility is  to always try to satisfy the safe-set constraint at the time steps $r-1$ and $N$. Our implementation of \emph{High} does it, therefore it ensures recursive feasibility.
\emph{Uniform} does not necessarily do it, but as long as the ratio between $N$ and the number of computational units $K$ is not large, \emph{Uniform} ensures recursive feasibility under a slightly stronger assumption. 
Indeed, \emph{Uniform} always tries to satisfy the safe-set constraint at a time step between $\lfloor N - N/K \rfloor$ and $N$.
This ensures recursive feasibility under the assumption that $\Vshat$ is $\lfloor N-L-N/K \rfloor$-step control invariant.
Finally, \emph{Closest} does not necessarily try to satisfy the safe-set constraint at time step $N$, therefore it does not ensure recursive feasibility.

While sharing the same theoretical properties, our tests show that \emph{Parallel-Constraint} MPC and its computationally cheaper variants in general outperform \emph{Receding-Constraint} MPC. 
In part this may be due to the use of the RTI scheme~\cite{diehl2005nominal}, which can lead to unfeasible solutions even when the tackled problem is feasible.


\section{Simulation Results}
\label{sec:results}
This section presents our simulation results\footnote{Our open-source code is freely available at \url{https://github.com/EliasFontanari/safe_mpc/tree/devel}.}.

\subsection{Simulation Setup}
To evaluate the proposed \emph{Parallel-Constraint} MPC, we have carried out a comparison of many MPC formulations:
\begin{itemize}
    \item \emph{Naive}: a classic formulation without terminal constraint, i.e.,  problem~\eqref{eq:mpc} with $\mathcal{X}_N = \Xs$. This is the baseline for all the experiments. 
    \item \emph{Receding}: the formulation~\eqref{eq:receding_constr_mpc}, using soft constraints for both $x_r\in\Vshat$ and $x_N\in\Vshat$. The penalty weight on the receding constraint ($10^4$) is higher than the terminal one ($w_s = 10^2$) to mimic a hard constraint.
    \item \emph{Parallel}: a formulation solving $N$ MPC problems in parallel, as described in Alg.~\ref{alg:parallel_mpc}.
    \item \emph{Uniform}, \emph{Closest}, \emph{High}: the three formulations of \emph{Parallel-Constraint} MPC using a limited number of computational units (either 4, 8, or 16), as described in Section~\ref{ssec:limited_computation}.
\end{itemize} 
For the simulations, we have considered a 3-joint manipulator, thus $n_x = 6,\, n_u = 3$. 
\textsc{CasADi}~\cite{Andersson2019} has been used as symbolic framework for the dynamics, costs, and constraints, while \textsc{Acados}~\cite{Verschueren2019} as OCPs solver and dynamics' integrator.
The task is a setpoint regulation problem with respect to a state purposely chosen near the position limit of the first joint, to test the safety of the controllers:
\begin{equation}
    x^{\text{ref}} = (q^{\text{max}} - 0.05, \bar q, \bar q, 0, 0, 0),
    \label{eq:target_value}
\end{equation}
with $\bar q = (q^{\text{max}}+q^{\text{min}})/2$.
We have used as running cost a least-squares function, penalizing deviations from $x^{\text{ref}}$ and  control efforts:
\begin{equation}
    \begin{aligned}
        l(x, u) &= ||x-x^\text{ref}||_Q^2 + ||u||_R^2, \\
        Q &= \text{diag}([500, 10^{-4} I_5]), \quad R = 10^{-4} I_3,
    \end{aligned}
    \label{eq:weight_values}
\end{equation}
where $I_h$ is the identity matrix with size $h$. 
Set membership to $\Vshat$ is verified with the constraint:
\begin{equation} \label{eq:terminal_constraint}
(1-\alpha) \phi(x) - ||\dot{q}|| \ge 0,     
\end{equation}
where $\phi(\cdot)$ is a Neural Network (NN) that computes an upper bound on the joint velocity norm~\cite{LaRocca2023}, and $\alpha \in [0, 1]$ is a safety margin that we introduced to ensure that $\Vshat \subseteq \Vs$. 
The PyTorch~\cite{pytorch} neural model is integrated inside \textsc{Acados} using the \textsc{L4CasADi}~\cite{salzmann2023learning} library.  
\begin{figure}[tbp]
    \centering
    \includegraphics[width=\columnwidth]{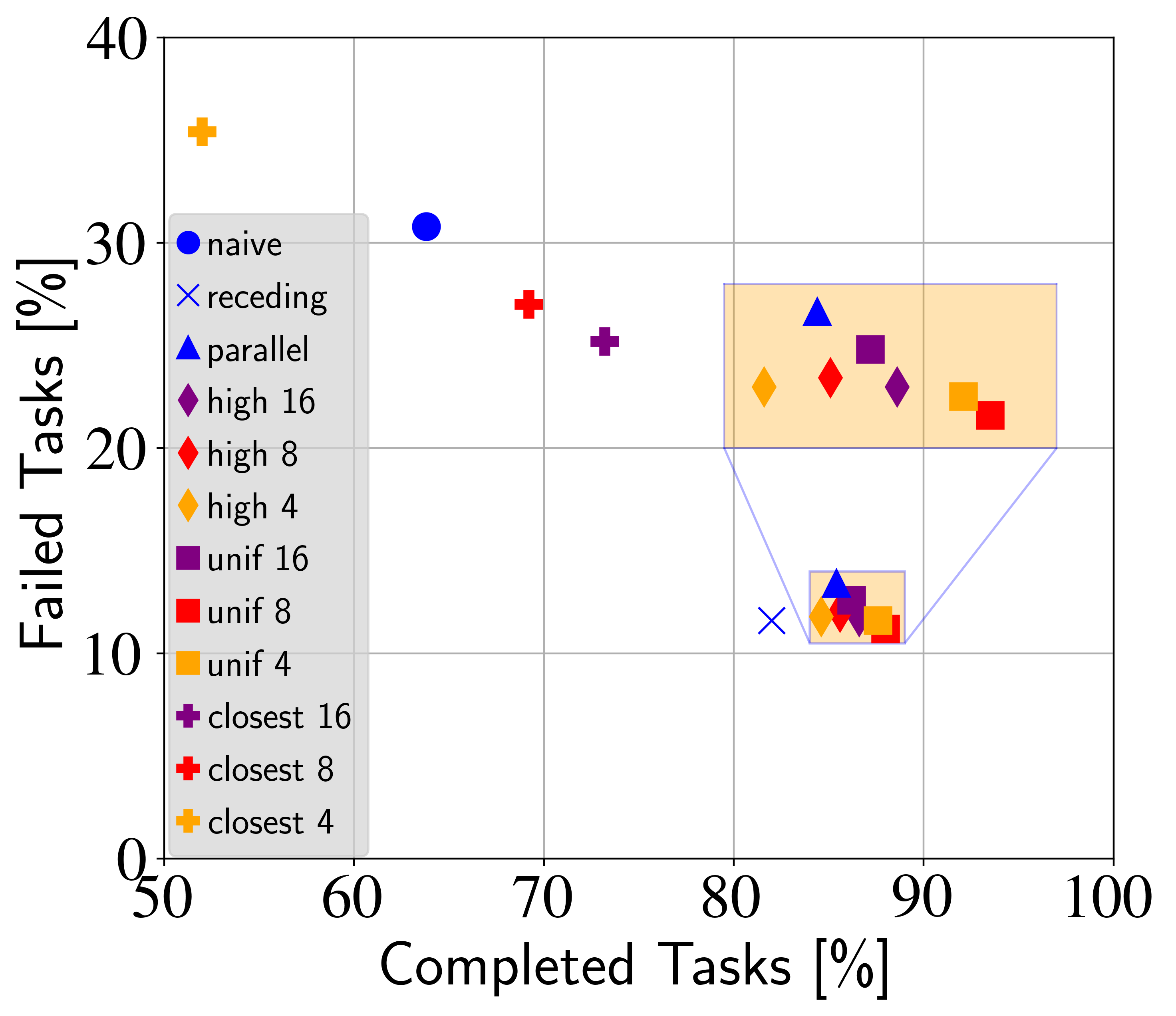}
    \caption{Number of times each controller completed the task, or violated a constraint (with safety factor $\alpha=2\%$, whose effect is reported in~\eqref{eq:terminal_constraint}). The orange rectangle is zoomed in to improve readability.}
    \label{fig:comparison_alpha_2}
\end{figure}
We have run 500 simulations for each MPC formulation, starting from the same 500 random joint positions $q_0$ with $\dot{q}_0 = 0$. The time step of the MPC was $dt = \SI{5}{\milli\second}$, and the horizon has been fixed to $N=35$, 
so that each iteration takes less than $\SI{4}{\milli\second}$ (leaving $\SI{1}{\milli\second}$ for further operations, to mimic the timing limitations of a real-time application). 
Each simulation was stopped when one of these conditions was met: i) the first joint has reached the target, ii) the robot has violated a joint limit, iii) the maximum number of simulation steps (600) has been reached.
In the last case, we have tried to trigger the safe abort from the final state, and considered the simulation as a failure if the safe abort has not worked. 

Optimistically we have set $L=0$, assuming that the safe-abort OCP can be solved in under $\SI{5}{\milli\second}$. 
While this is not currently the case, we hope we can achieve this performance in the future, for instance by training a neural network to provide a good warm-start for the solver.

\subsection{Discussion of the results}

\begin{figure}[tbp]
    \centering
    \includegraphics[width=\columnwidth]{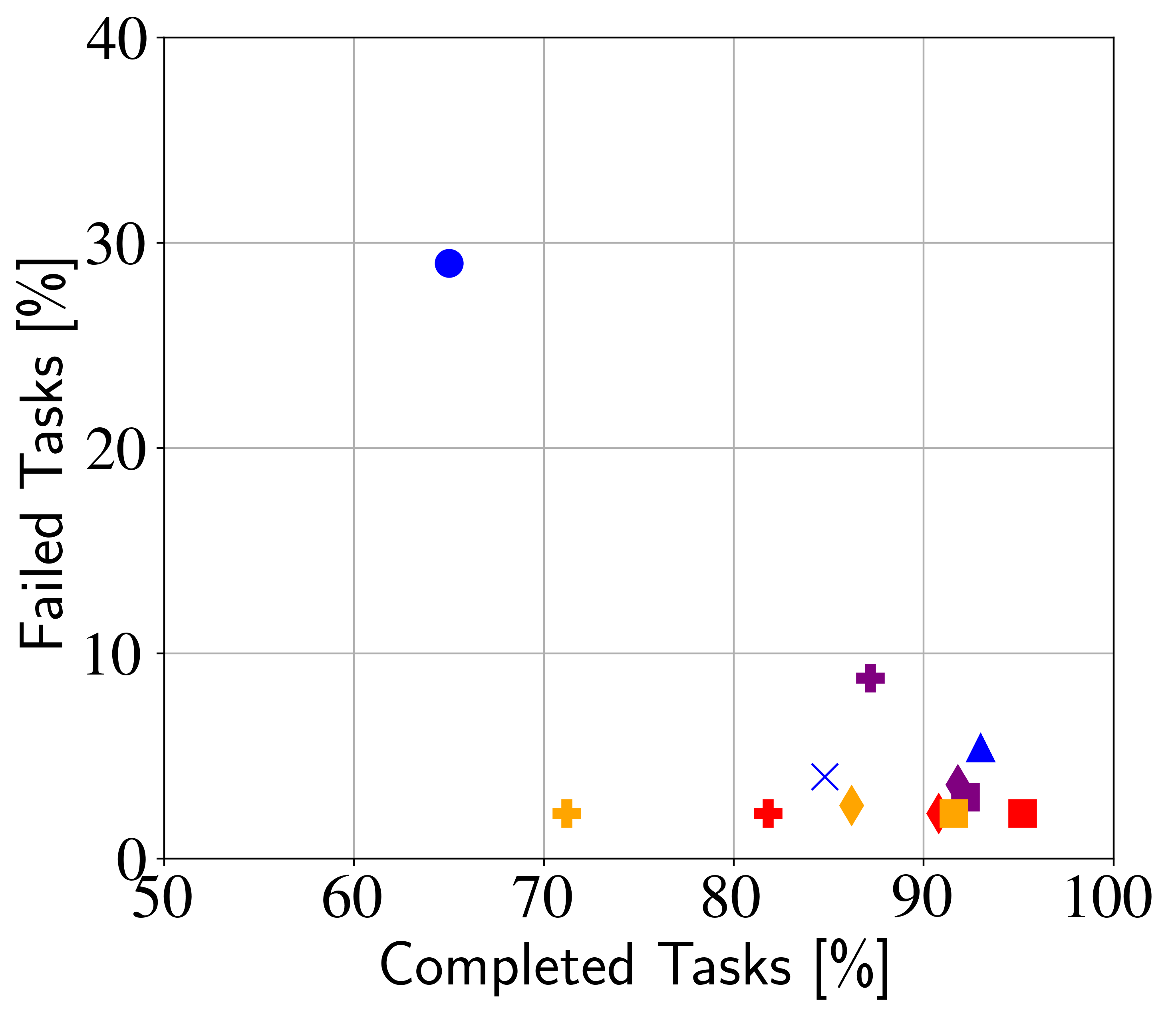}
    \caption{Number of times each controller completed the task, or violated a constraint (with $\alpha=10\%$).}
    \label{fig:comparison_alpha_10}
\end{figure}
\begin{figure}[tbp]
    \centering
    \includegraphics[width=\columnwidth]{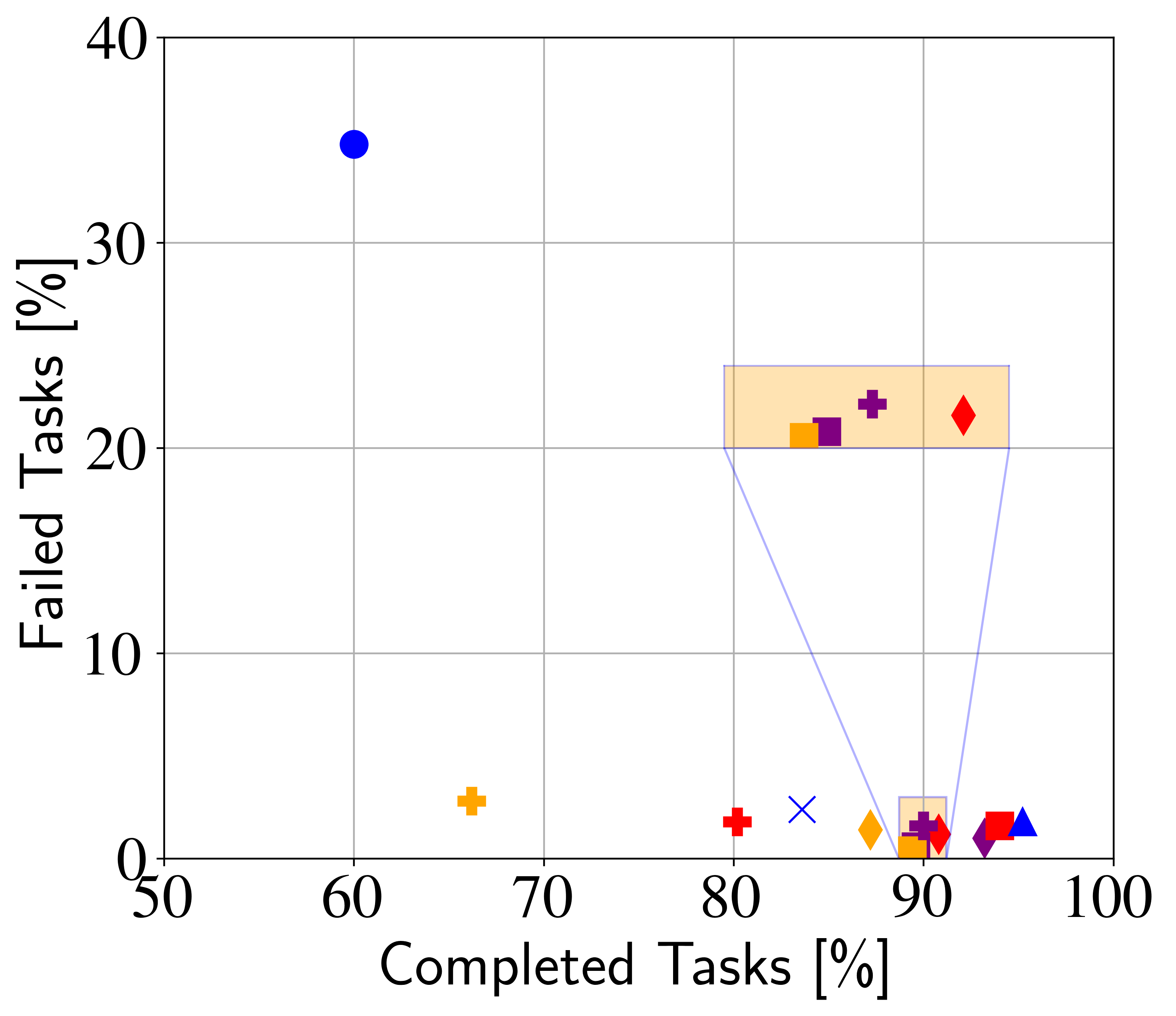}
    \caption{Number of times each controller completed the task, or violated a constraint (with $\alpha=15\%$).}
    \label{fig:comparison_alpha_15}
\end{figure}
Fig.~\ref{fig:comparison_alpha_2}, \ref{fig:comparison_alpha_10}, and~\ref{fig:comparison_alpha_15} summarize the results for the three tested scenarios, using $\alpha=2\%, 10\%$, and $15\%$, respectively.
For each value of $\alpha$ we have sampled a different set of initial conditions, making sure that the first MPC problem could be solved for every formulation.
The plots show the percentage of tasks that have been completed and failed by each MPC formulation. 
The percentage of tasks that have been safely aborted is not directly represented, but it can be retrieved knowing that the three values (completed, failed, aborted) must sum up to 100.

\subsubsection{Naive, Receding and Parallel}
For all values of $\alpha$, \emph{Naive} has been outperformed by \emph{Receding} and \emph{Parallel}, which have always completed more tasks while failing less tasks.
\emph{Parallel} and \emph{Receding} have performed somewhat similarly for $\alpha=2\%$: only $+4\%$ of completed tasks and $+1.8\%$ of failures for \emph{Parallel}. 
For higher values of $\alpha$ ($\alpha=10/15\%$), \emph{Parallel} has outperformed \emph{Receding} in terms of completed tasks ($+8.2/11.6\%$) while failing a similar number of tasks ($+1.4/-0.6\%$).
Surprisingly, the performance of \emph{Parallel} has monotonically improved increasing $\alpha$.
This was against our expectation, which was to witness a decrease in both the numbers of failed and completed tasks.
In particular, the most significant improvement is in the percentage of failed tasks, which went from $13.4\%$ (for $\alpha=2\%$) to $1.8\%$ (for $\alpha=15\%$).
This clearly shows that having a sufficiently high value of $\alpha$ is fundamental for getting a high success rate of the safe-abort procedure, which failed $91\%$ of the time for $\alpha=2\%$, but only $37\%$ of the time for $\alpha=15\%$.

\begin{figure}[tbp]
    \centering
    \includegraphics[width=\columnwidth]{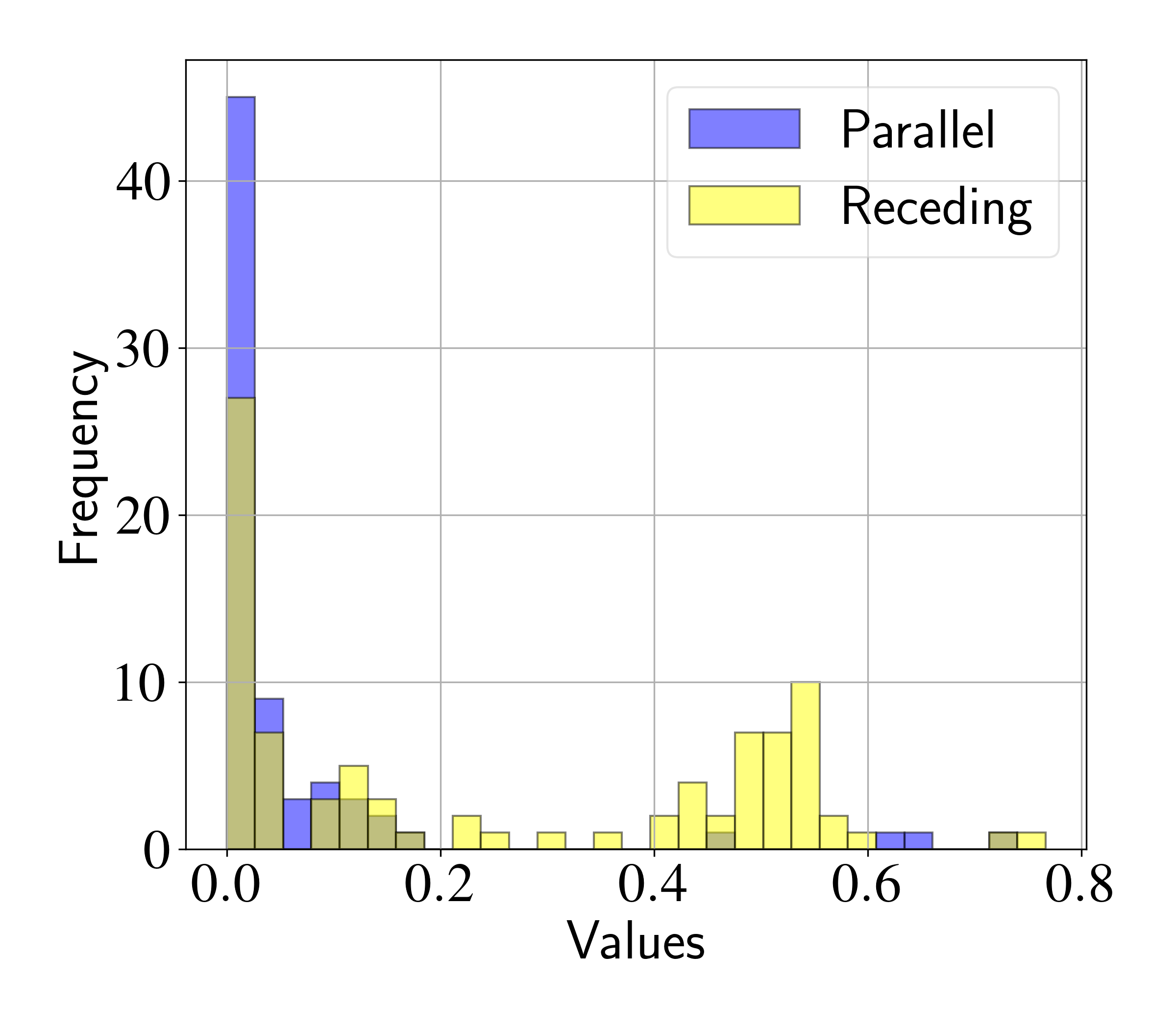}
    \caption{Distance to the boundary of the first joint when the safe-abort procedure is triggered (with $\alpha=2\%$).}
    \label{fig:joint_distance}
\end{figure}

We have noticed that for $\alpha=2\%$ the failure rate of the safe abort for \emph{Receding} was significantly lower than for \emph{Parallel} ($65\%$ against $91\%$). To understand why, we have analyzed the distance of the robot joints to the position boundaries at the states used for triggering the safe abort. 
In particular, we have noticed that the first joint was particularly crucial in this regard, so we have plotted the distance of this joint to its boundary, which can be seen in Fig.~\ref{fig:joint_distance}.
We can see how \emph{Parallel} has triggered the safe abort from states where the first joint was (on average) closer to its limit.
This can justify why the safe abort has failed more often for \emph{Parallel} than for \emph{Receding}.

We also analyzed the number of time steps taken to reach the target of \emph{Receding} and \emph{Parallel}, considering only the tasks completed by both controllers. \emph{Parallel} was faster to reach the target: on average it needed 225 steps, while \emph{Receding} required 262 steps.

\subsubsection{Limited Computational Units}
Regarding the other versions of \emph{Parallel} using a limited number of computational units, their performance has been mixed.
In general, \emph{High} and \emph{Unif} have performed well, approximately matching the performance of \emph{Parallel} when using 16 computational units, and showing a graceful performance degradation with lower numbers of computational units.
Even for 4 computational units, these strategies have always outperformed \emph{Receding}, empirically proving that improvements are achievable even with limited computational resources.
Surprisingly, for $\alpha=2\%$ and $\alpha=10\%$ some variants of \emph{High} and \emph{Unif} outperformed \emph{Parallel}; moreover \emph{Unif} has performed better when using 8 computational units rather than 16 or 4. 
These unexpected results could be statistical outliers due to the relatively small number of tests carried out (500), or they could be due to some phenomena that we have not yet understood.
%
\emph{Closest} instead has always been the worst performer, especially in terms of completed tasks (often worse than \emph{Receding}).

\section{Conclusions}
We have presented \emph{Parallel-Constraint} MPC, a novel MPC formulation that exploits parallel computation to improve safety.
Our formulation can be seen as an extension of our previous work on \emph{Receding-Constraint} MPC~\cite{Lunardi2024}, which also addressed the safety problem, but without exploiting parallel computation. 
While \emph{Parallel} and \emph{Receding-}Constraint MPC share similar theoretical properties in terms of safety and recursive feasibility, their empirical performance is substantially different. 
Our results show that exploiting parallel computation can lead to remarkable improvements. For instance, considering their best performance, \emph{Parallel} has completed $13\%$ more tasks than \emph{Receding}, while failing $5\%$ less tasks.
Moreover, we have shown that these improvements degrade gracefully when using a limited number of computational cores (4, 8 or 16) and that allocating these units using the right strategy is fundamental.

While the obtained results are highly encouraging, much work is still needed to allow the implementation of these advanced MPC formulations on real hardware. 
In the future, we would like to bridge the gap that is currently preventing us from validating our work on real robots. 
In particular, we are currently working on strategies to speed-up the safe-abort procedure, so that it is real-time compatible. 
Moreover, we are testing our methodology in more challenging scenarios featuring more complex constraints, such as avoiding self collisions and collisions with the environment.

Another aspect that we have not addressed yet is the presence of uncertainties. 
Exploiting tools from Robust MPC~\cite{Yu2013}, we could generate a nominal state trajectory using our approach (ignoring uncertainties) and then design a local control law to maintain the real system trajectory in a tube centered around the nominal trajectory~\cite{Mayne2011}.

\addtolength{\textheight}{-0cm}   


\newpage
\bibliographystyle{IEEEtran}
\bibliography{IEEEabrv,references}

\end{document}